\documentclass[final]{llncs}
\usepackage{makeidx}  
\usepackage{graphicx,booktabs,comment}
\usepackage[font=footnotesize]{caption}

\usepackage{amsmath}
\usepackage{mathtools}

\usepackage{amsmath}

\usepackage{amsfonts}

\usepackage{amssymb}

\usepackage{stmaryrd}

\usepackage[mathscr]{eucal}

\usepackage{amsbsy}

\usepackage{bm}

\usepackage{bbding}

\usepackage{array}

\usepackage[usenames]{color}

\usepackage{tabularx}

\usepackage{paralist}

\usepackage{fancyvrb}

\usepackage[position=below]{subfig}

\usepackage{boxedminipage}

\usepackage{multirow}
\usepackage{algorithmic}

\usepackage{graphicx}
\usepackage{ifpdf}
\ifpdf
\DeclareGraphicsExtensions{.pdf,.png}
\else
\DeclareGraphicsExtensions{.eps}
\fi

\usepackage[draft,hypertexnames=false]{hyperref}

\usepackage[notref,notcite]{showkeys}

\newcommand{\Sec}[1]{\hyperref[sec:#1]{\S\ref*{sec:#1}}} 
\newcommand{\Section}[1]{\hyperref[sec:#1]{Section~\ref*{sec:#1}}} 
\newcommand{\App}[1]{\hyperref[sec:#1]{\ref*{sec:#1}}} 
\newcommand{\Eqn}[1]{\hyperref[eq:#1]{(\ref*{eq:#1})}} 
\newcommand{\Fig}[1]{\hyperref[fig:#1]{Figure~\ref*{fig:#1}}} 
\newcommand{\Tab}[1]{\hyperref[tab:#1]{Table~\ref*{tab:#1}}} 
\newcommand{\Thm}[1]{\hyperref[thm:#1]{Theorem~\ref*{thm:#1}}} 
\newcommand{\Cor}[1]{\hyperref[cor:#1]{Corollary~\ref*{cor:#1}}} 
\newcommand{\Alg}[1]{\hyperref[alg:#1]{Algorithm~\ref*{alg:#1}}} 
\newcommand{\Def}[1]{\hyperref[def:#1]{Definition~\ref*{def:#1}}} 

\begin{document}
\frontmatter          
\pagestyle{headings}  
\addtocmark{Hamiltonian Mechanics} 
\title{Link Prediction via Generalized Coupled Tensor Factorisation}
\titlerunning{Link Prediction via Generalized Coupled Tensor Factorisation}  
%
\author{Beyza Ermis \inst{1}  \and Evrim Acar \inst{2} \and A. Taylan Cemgil\inst{1}}
\authorrunning{Beyza Ermis et al.} 

\tocauthor{Beyza Ermis, ...}
\institute{Bo\u{g}azi\c ci University\\ 34342, Bebek, Istanbul, Turkey,\\
\email{beyza.ermis@boun.edu.tr,taylan.cemgil@boun.edu.tr},
\and
University of Copenhagen\\ DK-1958 Frederiksberg C, Denmark\\
\email{evrim@life.ku.dk}}

\maketitle              

\begin{abstract}

This study deals with the \emph{missing link prediction} problem: the problem of predicting the existence of missing connections between entities of interest. We address link prediction using coupled analysis of relational datasets represented as heterogeneous data, i.e., datasets in the form of matrices and higher-order tensors. 
We propose to use an approach based on probabilistic interpretation of tensor factorisation models, i.e., 
Generalised Coupled Tensor Factorisation, which can simultaneously fit \emph{a large class of tensor models} to higher-order tensors/matrices with common latent factors using \emph{different loss functions}. 
Numerical experiments demonstrate that joint analysis of data from multiple sources via coupled factorisation improves the link prediction performance and the selection of right loss function and tensor model is crucial for accurately predicting missing links.

\end{abstract}


\section{Introduction}
\label{sec:intro}

Recent technological advances, such as the Internet,  multi-media devices or social networks provide abundance of relational data. For instance, in retail recommender systems, in addition to retail data showing \emph{who has bought which items}, we may also have access to customers' social networks, i.e., \emph{who is friends with whom}. In such complex problems, jointly analyzing data from multiple sources has great potential to increase our ability for capturing the underlying structure in data. Data fusion, therefore, is a viable candidate for addressing the challenging link prediction problem. Applications in many areas including recommender systems and social network analysis deal with link prediction, i.e., the problem of inferring whether there is a relation between the entities of interest. 
For instance, if a customer buys an item, the customer and the item can be considered to be linked. The task of recommending other items the customer may be interested in can be cast as a missing link prediction problem. However, the results are likely to be poor if the prediction is done in isolation on a single view of data. Such datasets, whilst large in dimension, are already very sparse \cite{GeDi05} and potentially represent only a very incomplete picture of the reality \cite{ClMoNe08}. Therefore, relational data from other sources is often incorporated into link prediction models \cite{HaZa11}.

Matrix factorisations have proved to be very useful in recommender systems \cite{KoBeVo09}. An effective way of including side information via additional relational data in a link prediction model is to represent different relations as a collection of matrices.  Subsequently, this collection of matrices are jointly analyzed using collective matrix factorisation \cite{LoZhWuYu06,SiGo08a}. In many applications, however, matrices are not sufficient for a faithful representation of multiple attributes, and higher-order tensor and matrix factorisation methods are needed. An influential study in this direction is by Banerjee et al. \cite{BaBaMe07}, where a general clustering method for joint analysis of heterogeneous data has been studied. The goal here is clustering entities based on multiple relations, where each relation is represented as a matrix (e.g., movies by review words matrix showing movie reviews) or a higher-order tensor (e.g., movies by viewers by actors tensor showing viewers' ratings). 

\begin{figure}[t!]
\centering
\includegraphics[width=0.35\textwidth,trim=0 0 0 0,clip]{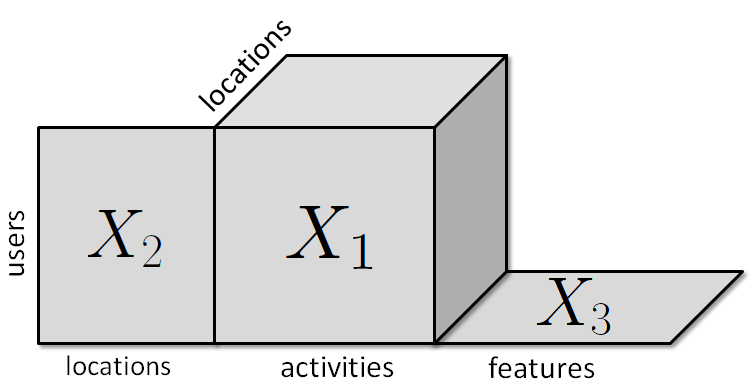}
\caption{A third-order tensor coupled with two matrices in two different modes.}
\vspace{-20pt}
\label{fig:Data}
\end{figure}

In this paper, we address link prediction problem using coupled analysis of datasets in the form of matrices and higher-order tensors. As an example application, we study a real-world GPS (Global Positioning System) dataset \cite{ZhCaZhXiYa10} for location-activity recommendation such that given an incomplete dataset showing which users perform which activities at various locations, we would like to fill in the missing links between (user, activity, location) triplets ($X_1$). We also make use of additional sources of information showing the locations visited by users based on GPS trajectories ($X_2$) and the features of locations in terms of number of different points of interest at each location ($X_3$) (\Fig{Data}).

Various algorithms have been proposed in the literature for coupled analysis of heterogeneous data. Lin et al. \cite{LiSuCaKo09} addresses the community extraction problem on multi-relational data using a coupled factorisation approach modeling higher-order tensors using a specific tensor model, i.e., CANDECOMP/PARAFAC (CP) \cite{Ha70,CaCh70}, and a Kullback-Leibler (KL) divergence-based cost function. Also, a recent study by Narita et al. \cite{NaHaToKa11} has addressed the tensor completion problem using additional data using a Euclidean distance-based loss function. Unlike previous studies, we use an approach, i.e., Generalized Coupled Tensor Factorisations (GCTF) \cite{yilmazGTF}, which enables us to investigate alternative tensor models and cost functions for coupled analysis of heterogeneous data. The main contributions of this paper can be summarized as follows:
\begin{itemize}
\item We consider different tensor models, i.e., CP and Tucker \cite{Tu66}, and loss functions, i.e., KL-divergence and Euclidean distance, for joint analysis of heterogeneous data for link prediction using the GCTF framework.
\item Using a real GPS data set, we demonstrate that coupled tensor factorisations outperform low-rank approximations of a single tensor in terms of missing link prediction and the selection of the tensor model as well as the loss function is significant in terms of link prediction performance.
\item We also demonstrate that it is possible to address the cold-start problem in link prediction using the proposed coupled models.
\end{itemize}


The rest of the paper is organized as follows. In \Sec{related}, we survey the related work on link prediction as well as joint factorisation of data. \Sec{method} introduces our algorithmic framework, i.e., GCTF, while \Sec{link_prediction} discusses its adaptation for the link prediction problem. Experimental results on a real dataset are presented in \Sec{exp}. Finally, we conclude with future work in \Sec{conc}.

\section{Related Work}
\label{sec:related}

In order to deal with the challenging task of link prediction, many studies have proposed to exploit multi-relational nature of the data and showed improved link prediction performance by incorporating related sources of information in their modeling framework. For instance, Taskar et al \cite{TaWoAbKo03} use relational Markov networks that model links between entities as well as their attributes. Popescul and Ungan \cite{PoUn03} extract relational features to learn the existence of links (see \cite{HaZa11} for a comprehensive list of similar studies).

For analysis of multi-relational data, Singh and Gordon \cite{SiGo08a} as well as Long et al. \cite{LoZhWuYu06} introduce collective matrix factorisation. Matrix factorisation-based techniques have proved useful in terms of capturing the underlying patterns in data, e.g., in recommender systems \cite{KoBeVo09}, and joint analysis of matrices has been widely applied in numerous disciplines including signal processing \cite{YoKiKaCh10} and bioinformatics \cite{AlBrBo03}. Recent studies extend collective matrix factorisation to coupled analysis of multi-relational data in the form of matrices and higher-order tensors \cite{SmWeBo00,BaBaMe07} since in many disciplines, relations can be defined among more than two entities, e.g., when a user engages in an activity at a certain location, a relation can be defined over user, activity and location entities.  Banerjee et al. \cite{BaBaMe07} introduced a multi-way clustering approach for relational and multi-relational data where coupled analysis of heterogeneous data was studied using minimum Bregman information. Lin et al. \cite{LiSuCaKo09} also discussed coupled matrix and tensor factorisations using KL-divergence modeling higher-order tensors by fitting a CP model. While these studies use alternating algorithms, Acar et al. \cite{AcKoDu11b} proposed an all-at-once optimization approach for coupled analysis.

Missing link prediction is also closely related to matrix and tensor completion studies. By using a low-rank structure of a data set, it is possible to recover missing entries for matrices \cite{CaPl09} and higher-order tensors \cite{AcDuKoMo10}.

Note that we focus on missing link prediction in this paper and do not address the temporal link prediction problem, where snapshots of the set of links up to time $t$ are given and the goal is to predict the links at time $t + 1$. Tensor factorisations have previously been used for temporal link prediction \cite{AcDuKo10b} but using only a single source of information. 



\section{Methodology}
\label{sec:method}

In this study, we discuss Generalized Coupled Tensor Factorisation framework\cite{yilmazGTF} for coupled factorisation of several tensors and matrices to fill in the missing links in observed data.
A generalized tensor factorisation problem is specified by an observed tensor $X$ (with possibly missing entries) and a collection of latent tensors to be estimated, $Z_{1:|\alpha|} = \lbrace Z_\alpha \rbrace$ for $\alpha = 1 . . . |\alpha|$. 

GCTF framework is a generalisation of the Probabilistic Latent Tensor Factorisation (PLTF) \cite{Yilmaz:2010} to coupled factorisation. 
In this framework, the goal is to compute an approximate factorisation of $X$ in terms of a product of individual factors $Z_\alpha$. Here, we define $V$ as the set of all indices in a model, $V_0$ as the set of visible indices, $V_\alpha$ as the set of indices in $Z_\alpha$, and $\bar{V}_\alpha = V- V_\alpha$ as the set of all indices not in $Z_\alpha$. We use small letters as $v_\alpha$ to refer to a particular setting of indices in $V_\alpha$.

PLTF tries to solve the following approximation problem
\small
\begin{align}
	X(v_0) \approx \hat{X}(v_0)=\sum_{\bar{v}_0} \prod_\alpha Z_\alpha(v_\alpha), \label{eqn:tensorFact}
\end{align}
\normalsize

Since the product $\prod_\alpha Z_\alpha(v_\alpha)$ is collapsed over a set of indices, the factorisation is latent. The approximation problem is cast as an optimisation problem minimizing the divergence $d(X, \hat{X})$, where $d$ is a divergence (a quasi-squared-distance) between the observed data $X$ and model prediction $\hat{X}$. In applications, $d$ is typically Euclidean (EUC), Kullback-Leibler (KL) or Itakura-Saito (IS) \cite{yilmazGTF}.

In this paper, we use non-negative variants of the two most widely-used low-rank tensor factorisation models, i.e., Tucker model, and the more restricted CP model, as baseline methods in \Sec{exp}. These models can be defined in the PLTF notation as follows. Given a three-way tensor $X$, its CP model is defined as:
\small
\begin{align}
    X(i,j,k) \approx \hat{X}(i,j,k)  &= \sum_r Z_1(i,r) Z_2(j,r) Z_3(k,r) \label{eq:CP}
\end{align}
\normalsize

where index sets $V=\{i,j,k,r\}$, $V_0=\{i,j,k\}$, $V_1=\{i,r\}$, $V_2=\{j,r\}$ and $V_3=\{k,r\}$.
A Tucker model of $X$ is defined in the PLTF notation as follows:
\small
\begin{align}
    X(i,j,k) \approx \hat{X}(i,j,k)  &= \sum_{p,q,r} Z_1(i,p) Z_2(j,q) Z_3(k,r) Z_4(p,q,r) \label{eq:Tucker}
\end{align}
\normalsize

where index sets $V=\{i,j,k,p,q,r\}$, $V_0=\{i,j,k\}$, $V_1=\{i,p\}$, $V_2=\{j,q\}$, $V_3=\{k,r\}$ and $V_4=\{p,q,r\} $.

The update equation for non-negative generalized tensor factorisation can be used for both (\ref{eq:CP}) and (\ref{eq:Tucker}) and is expressed as:
\small
\begin{align}
	Z_\alpha \leftarrow Z_\alpha \circ \frac{{\Delta_{\alpha} (M \circ \hat{X}^{-p} \circ X) }} { { \Delta_{\alpha} (M \circ \hat{X}^{1-p} ) }}
	\hspace{10mm} s.t. \hspace{3mm}   Z_\alpha(v_{\alpha}) > 0.
	\label{eqn:updateeq1}
\end{align}
\normalsize
where $\circ$ is the Hadamard product (element-wise product), $M$ is a $0-1$ mask array with  $M(v_{0})=1$ ($M(v_{0})=0$) if $X(v_{0})$ is observed (missing). Here $p$ determines the cost function, i.e., $p=\{0,1,2\}$ correspond to the $\beta$-divergence \cite{cichocki09} that unifies EUC, KL, and IS cost functions, respectively. In this iteration, we define the tensor valued function $\Delta_{\alpha}(A)$ as:
\small
\begin{align}
   \Delta_{\alpha}(A) = \sum_{\bar{v}_\alpha}  A(v_{0} )  \prod_{\alpha'\neq\alpha} {Z_{\alpha'}(v_{\alpha'})} 
\end{align}
\normalsize
$\Delta_{\alpha}(A)$ is an object, the same size of $Z_\alpha$, obtained simply by multiplying all factors other than the one being updated with an object of the order of the data. Hence the key observation is that the $\Delta_{\alpha}$ function is just computing a tensor product and collapses this product over indices not appearing in $Z_\alpha$, which is algebraically equivalent to computing a marginal sum.

As an example, for KL cost, we rewrite (\ref{eqn:updateeq1}) more compactly as:
\small
\begin{align}
   Z_\alpha \leftarrow Z_\alpha \circ \Delta_{\alpha}(M \circ X / \hat{X}) / \Delta_{\alpha}(M)
\end{align}
\normalsize
This update rule can be used iteratively for all non-negative $Z_\alpha$ and converges to a local minimum provided we start from some non-negative initial values. 
For updating $Z_\alpha$, we need to compute the $\Delta$ function twice for arguments $A = M_\nu \circ \hat{X}_\nu^{-p} \circ X_\nu $ and $A=M_\nu \circ \hat{X}_\nu^{1-p}$. 

\subsection{Generalized Coupled Tensor Factorisation}
The Generalised Coupled Tensor Factorisation model takes the PLTF model one step further where, we have multiple observed tensors $X_\nu$ that are factorised simultaneously:
\small
\begin{align}
	X_\nu(v_{0,\nu}) \approx \hat{X}_\nu(v_{0,\nu}) = \sum_{\bar{v}_{0,\nu}}  \prod_\alpha Z_\alpha(v_\alpha)^{R^{\nu,\alpha}} \label{eqn:gctf}
\end{align}
\normalsize
where $\nu = 1,... |\nu|$ and $R$ is a {\it coupling matrix} defined as follows:
\small
 \begin{align}
  R^{\nu,\alpha} &= \left\{
   \begin{array}{l l}
      1 &  \quad \text{if $X_\nu$ and $Z_\alpha$ connected}\\
      0 &  \quad \text{otherwise}  \\
   \end{array} \right. .
\end{align}
\normalsize
Note that, distinct from PLTF model, there are multiple visible index sets ($V_{0,\nu}$) in the GCTF model, each specifying the attributes of the observed tensor $X_\nu$.

The inference, i.e., estimation of the shared latent factors $Z_\alpha$, can be achieved via iterative optimisation (see \cite{yilmazGTF}). For non-negative data and factors, one can obtain the following compact fixed point equation where each $Z_\alpha$ is updated in an alternating fashion fixing the other factors $Z_{\alpha'}$, for $\alpha' \neq \alpha$:
\small
\begin{align}
	Z_\alpha \leftarrow Z_\alpha \circ \frac{{\sum_\nu R^{\nu,\alpha} \Delta_{\alpha,\nu}(M_\nu \circ \hat{X}_\nu^{-p} \circ X_\nu  ) }} { {  \sum_\nu R^{\nu,\alpha}  \Delta_{\alpha,\nu} (M_\nu \circ \hat{X}_\nu^{1-p} )    }}.
	\label{eqn:updateeq}
\end{align}
\normalsize
where $M_\nu$ is a $0-1$ mask array with  $M_\nu(v_{0,\nu})=1$ ($M_\nu(v_{0,\nu})=0$) if $X_\nu(v_{0,\nu})$ is observed (missing). Here $p$, as in (\ref{eqn:updateeq1}), determines the cost function, i.e., $p=\{0,1\}$ corresponds to EUC and KL cost functions, respectively (see Table~\ref{table:update_rules}). In this iteration, the key quantity is the $\Delta_{\alpha,\nu}$ function defined as follows:
\small
\begin{align}
   \Delta_{\alpha,\nu}(A) = \left[ \sum_{v_{0,\nu} \cap \bar{v}_\alpha}   A(v_{0,\nu} )
   \sum_{\bar{v}_0 \cap \bar{v}_\alpha} \prod_{\alpha'\neq\alpha} {Z_{\alpha'}(v_{\alpha'})}^{R^{\nu,\alpha'}} \right]
\end{align}
\normalsize

 \begin{table}[t]
		\centering
		\caption{Update rules for different $p$ values }
		\begin{tabular}{l c c}
		    \toprule
		$p$	  & Cost Function & Multiplicative Update Rule \\
	\midrule
	$0$  & Euclidean & $Z_\alpha \leftarrow Z_\alpha \circ \frac{{\sum_\nu R^{\nu,\alpha} \Delta_{\alpha,\nu}(M_\nu  \circ X_\nu  ) }} { {  \sum_\nu R^{\nu,\alpha}  \Delta_{\alpha,\nu} (M_\nu \circ \hat{X}_\nu )    }}$\\
	\midrule
	$1$  & Kullback-Leibler    &$Z_\alpha \leftarrow Z_\alpha \circ \frac{{\sum_\nu R^{\nu,\alpha} \Delta_{\alpha,\nu}(M_\nu \circ \hat{X}_\nu^{-1} \circ X_\nu  ) }} { {  \sum_\nu R^{\nu,\alpha}  \Delta_{\alpha,\nu} (M_\nu )    }}$  \\
	\bottomrule
	\end{tabular}
	\label{table:update_rules}
	\vspace{-15pt}
\end{table}

Assuming that all datasets have equal number of dimensions, i..e, a tensor is an $N \times N \times N$ array while the coupled matrix is of size $N \times N$, then the leading term in the computational complexity of the coupled model will be due to the updates for the tensor model. For an $F$-component CP model, for instance, that would be $O(N^{3}F)$. The updates can be implemented by taking into account the sparsity pattern of the data.
\section{Link Prediction with Coupled Tensor Factorisation}
\label{sec:link_prediction}

In this section, by using the GCTF framework, we propose a solution for link prediction task with different coupled models and loss functions. 
We have a three-way observation tensor ${X}_1$ with elements $0$ and $1$, where $0$ denotes a known absent link and $1$ denotes a known present link, and two auxiliary matrices ${X}_2$ and ${X}_3$ that provide side information. Our aim is to restore the missing links in ${X}_1$. This is a difficult link prediction problem since ${X}_1$ contains less than 1\% of all possible links or an entire slice of ${X}_1$ may be missing. Using low-rank factorisation of a tensor to estimate missing entries will be ineffective, in particular, in the case of structured missing data such as missing slices. 
In terms of coupled models, we are not restricted to a specific model topology, i.e., since we use the GCTF framework, we can design application-specific models. The choice of a particular factorisation model is strongly guided by the application; therefore, we first give a brief description of the data set. 

UCLAF dataset\footnote{\url{http://www.cse.ust.hk/~vincentz/aaai10.uclaf.data.mat}} \cite{ZhCaZhXiYa10} is extracted from the GPS data that include information of three types of entities: user, location and activity. The relations between user-location-activity triplets are used to construct a three-way tensor $X_{1}$. In tensor $X_{1}$, an entry $X_{1}(i,j,k)$ indicates the frequency of a user $i$ visiting location $j$ and doing activity $k$ there; otherwise, it is $0$. Since we address the link prediction problem in this study, we define the user-location-activity tensor $X_{1}$ as:
\footnotesize
 \begin{align*}
  X_{1}(i,j,k) &= \left\{
   \begin{array}{l l}
      1 & \quad \text{if user $i$ visits location $j$ and performs activity $k$ there} \\
      0 & \quad \text{otherwise}
   \end{array} \right. .
\end{align*}
\normalsize

The dataset has been constructed by clustering raw GPS points into 168 meaningful locations and manually parsing the user comments attached to the GPS data into activity annotations for the 168 locations. Consequently, the data consists of 164 users, 168 locations and 5 different types of activities. (See   \cite{ZhCaZhXiYa10} for details).

Additionally, the collected data includes side information: the location features from the POI (points of interest) database as well as the user-location preferences from the GPS trajectory data, represented by matrix $X_{2}$ and $X_{3}$ respectively. In our model, the user-location preferences matrix of size $I \times J$ has entries $X_{2}(i, m)$, where $I$ is the number of users and $J$ is the number of locations and we use index $m$ as the location index instead of $j$. The rationale behind this choice is to relax the model as the entries in $X_{1}$ and $X_{2}$ are measuring distinct quantities: $X_{2}(i,m)$ represents the frequency of a user $i$ visiting location $m$ and stayed there over a time threshold while $X_1$ only indicates an activity by a specific user $i$ at location $j$.  The relation between the location entries $j$ and $m$ in $X_{1}$ and $X_{2}$ are coupled via a common factor over the users. Finally, we represent the location-feature values with matrix $X_{3}$ of size $J \times N$, where $J$ is the number of locations, that has the same location type in $X_{1}$, and $N$ is the number of features. In particular, an entry $X_{3}(j,n)$ represents the number of different POIs at location $j$. Using the location features, we could gain information about location similarities based on their feature values.

In this data set, 18 users have no location and activity information. Therefore, we have used the remaining 146 users. In order to decrease the effect of outliers,  location-feature matrix is preprocessed as follows: $ X_{3}(j,n) = 1 + \log(X_{3}(j,n))$ if $X_{3}(j,n) > 0$; otherwise,   $ X_{3}(j,n) =0$. In our experiments, number of users is $I=146$, number of locations $J=168$, number of activities $K=5$ and number of location features $N=14$.

We form two coupled models to fill in the missing links in tensor ${X}_1$. For both models, we use KL divergence and Euclidean as the cost functions in our non-negative decomposition problems. In the first model, we applied the coupled approach to a CP-style tensor model by analysing the tensor ${X}_1$ jointly with the additional matrices ${X}_2$ and ${X}_3$. This gives us the following model:
\footnotesize
\begin{align}\label{eqn:CPcoupled}
	\hat{X}_1({i,j,k}) &= \sum_r A({i,r}) B({j,r}) C({k,r})  \\ 
	\hat{X}_2({i,m}) &= \sum_r  A({i,r}) D({m,r}) \\
	\hat{X}_3({j,n}) &= \sum_r   B({j,r}) E({n,r}) 
\end{align}
\normalsize
Here, we have three observed tensors, that share common factors; therefore, we have a coupled tensor factorisation problem. The coupling matrix $R$ with $|\alpha|=5$, $|\nu|=3$ for this model is defined as follows:
\small
\begin{align}
  R &= \left[
   \begin{array}{l l l l l}
      1 & 1 & 1 & 0 & 0\\
      1 & 0 & 0 & 1 & 0\\
      0 & 1 & 0 & 0 & 1\\
   \end{array} \right]
   &&   \text{with }
   \begin{array}{l }
     \hat{X}_1 = \sum A^1 B^1 C^1 D^0 E^0 \\
     \hat{X}_2 = \sum A^1 B^0 C^0 D^1 E^0 \\
     \hat{X}_3 = \sum A^0 B^1 C^0 D^0 E^1 \\
   \end{array}.
\end{align}
\normalsize
Note that, ${X}_1$ and ${X}_2$ share the common factor matrix $A$ with entries $A(i, r)$;  we can interpret each row of $A(i,:)$ as user $i$'s latent position in a $|r|$ dimensional `preferences' space. The factor matrix $B$ with entries $B(j,r)$ represents the 
latent position of the location $j$ in the same preferences space. The user $i$ at location $j$ tends to perform activity $k$ where the weight $A(i,r) B(j,r)$ is large for at least one $r$, i.e., there is a match between the users preference and what the location `has to offer'. The location specific factor $B$ is also influenced by the location-feature matrix ${X}_3$ .
	
Following the same line of thought, we apply the coupled approach using a Tucker decomposition to form our second model, which is as follows:
\footnotesize
\begin{align}\label{eqn:TUCcoupled}  
  \hat{X}_1({i,j,k}) &= \sum_{p,_q,_r} A({i,p}) B({j,q}) C({k,r}) D({p,q,r}) \\
  \hat{X}_2({i,m}) &= \sum_p A({i,p}) E({m,p}) & \\
  \hat{X}_3({j,n}) &= \sum_r  B({j,q}) F({n,q})
  \vspace{-15pt}
\end{align}
\normalsize
In this model, once again, the factor $A$ is shared by ${X}_1$ and ${X}_2$, while the factor $B$ is shared by ${X}_1$ and ${X}_3$. In contrast to the coupled CP model in (\ref{eqn:CPcoupled}), this model assumes that user $i$ at location $j$ tends to perform activity $k$ with the weight $\sum_{p,q} A({i,p}) B({j,q}) D({p,q,r})$. Here, a latent preference space interpretation is less intuitive but the model has more freedom to represent the dependence.

\section{Experimental Results}
\label{sec:exp}

In this section, we assess the performance of the coupled models proposed in the previous section in terms of missing link prediction. First, we demonstrate that coupled tensor factorisations outperform low-rank approximations of a single tensor in terms of missing link prediction. Then we compare different tensor models and loss functions and show that selection of the tensor model and loss function is significant in terms of link prediction performance, especially when the fraction of unobserved elements is high. Furthermore, we study the case with completely missing slices, which corresponds to the cold-start problem in our link prediction setting and demonstrate that it is still possible to predict missing links using the proposed coupled models.  
\subsection{Experimental Setting}
We design experiments to evaluate the performance of our models in terms of link prediction. By setting different amounts of data to missing in user-location-activity tensor $X_{1}$, we compare the following models using both KL-divergence and the Euclidean as cost functions:
\begin{itemize}
\item \emph{Low-rank approximations of a single tensor:} (i) CP and  (ii) Tucker factorisation of user-location-activity tensor $X_{1}$,
\item \emph{Coupled tensor factorisations:} (i) CP factorisation of $X_{1}$ coupled with factorisation of user-location matrix $X_{2}$ and location-feature matrix $X_{3}$  (ii) Tucker factorisation of $X_{1}$ coupled with factorisation of $X_{2}$ and $X_{3}$.
\end{itemize}

We use two missing data patterns: (i) randomly missing entries, (ii) randomly missing slices. In all experiments, number of components, i.e., number of columns in each factor matrix, $Z_i$, is set to 2. To measure the link prediction performance, we use AUC (Area Under the Receiver Operating Characteristic Curve). 

\subsection{Results}		

\begin{figure}[h!]
\vspace{-20pt}
\begin{minipage}[b]{0.50\linewidth}
\centering
\subfloat[EUC with 80\% missing]{
\includegraphics[scale=0.35]{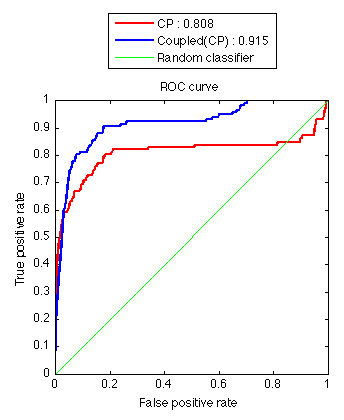}\label{fig:subfig1}}
\end{minipage}
\begin{minipage}[b]{0.50\linewidth}
\centering
\subfloat[KL with 80\% missing]{
\includegraphics[scale=0.35]{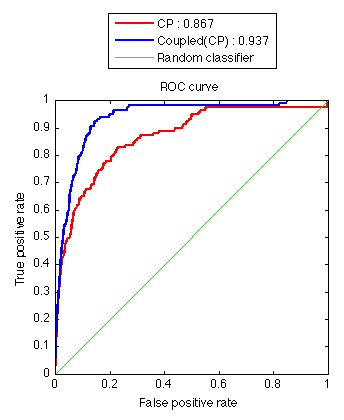}\label{fig:subfig2}}
\end{minipage}
\caption{Comparison of CP and Coupled(CP) models}
\label{fig:comp2modCP}
\vspace{-15pt}
\end{figure}

In order to demonstrate the power of coupled analysis, we compared the link prediction performance of standard CP and Tucker models with coupled ones using EUC and KL cost functions at different amounts, i.e., $\lbrace 40, 60, 80, 90, 95 \rbrace$, of randomly unobserved elements. For all cases, coupled models outperform the standard models clearly. Figure~\ref{fig:comp2modCP} shows the comparison of CP and coupled CP models with different cost functions when 80\% of the data is missing. As we can see, coupled models using additional information perform better than the standard models; in particular, when the percentage of missing data is high. When the fraction of missing data was more than 80\%, the standard models could not find a solution.

\begin{figure}[h!]
\vspace{-20pt}
\begin{minipage}[b]{0.50\linewidth}
\centering
\subfloat[Coupled (CP) Model]{
\includegraphics[scale=0.35]{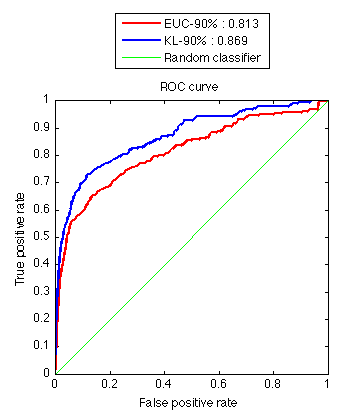}\label{fig:subfig7}}
\end{minipage}
\begin{minipage}[b]{0.50\linewidth}
\centering
\subfloat[Coupled (Tucker) Model]{
\includegraphics[scale=0.35]{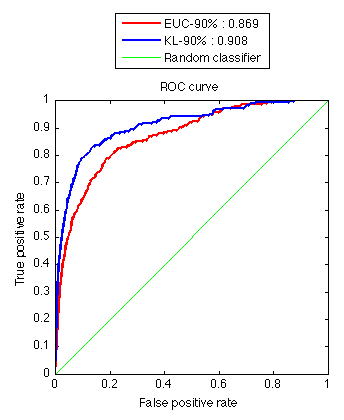}\label{fig:subfig8}}
\end{minipage}
\caption{Comparison of EUC distance and KL divergence with 90\% missing data}
\label{fig:EUCvsKL90}
\vspace{-15pt}
\end{figure}

In order to demonstrate the effect of the cost function modeling the data, we have also carried out experiments on both coupled CP and Tucker models at different missing data fractions. For all cases, the KL cost function seems to perform better than EUC, especially when the fraction of missing entries is high. Figure~\ref{fig:EUCvsKL90} illustrates the performance of Euclidean distance and Kullback-Leibler divergence for both coupled CP and Tucker models when 90\% of the data is unobserved.

\begin{figure}[h!]
\vspace{15pt}
\begin{minipage}[b]{0.50\linewidth}
\centering
\subfloat[CP vs Tucker - 40\% missing]{
\includegraphics[scale=0.35]{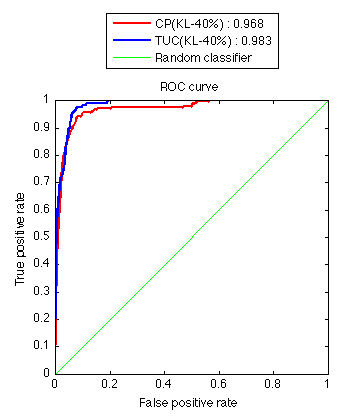}\label{fig:subfig10}}
\end{minipage}
\begin{minipage}[b]{0.50\linewidth}
\centering
\subfloat[CP vs Tucker - 90\% missing]{
\includegraphics[scale=0.35]{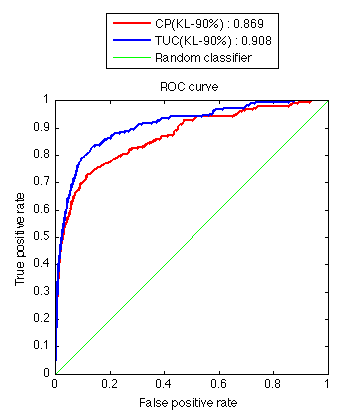}\label{fig:subfig12}}
\end{minipage}
\caption{Comparison of Coupled CP and Tucker models with KL}
\label{fig:CPvsTUC}
\vspace{-15pt}
\end{figure}

Finally, Figure~\ref{fig:CPvsTUC} shows the comparison of coupled CP and Tucker models in order to illustrate the tensor model modeling the data best. We can see that Tucker model outperforms the CP model; because Tucker model is more flexible due to the full core tensor which is helpful for us to explore the structural information embedded in the data. 

\subsubsection{Cold Start Problem:} 
\begin{figure}[h!]
\vspace{-20pt}
\begin{minipage}[b]{0.50\linewidth}
\centering
\subfloat[CP and Tucker - 10MS]{
\includegraphics[scale=0.35]{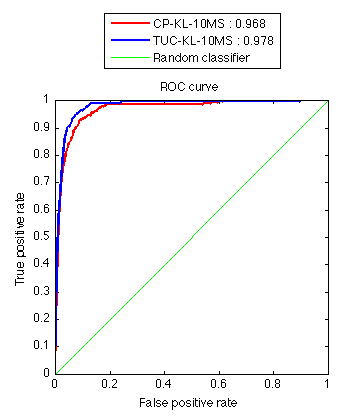}\label{fig:subfig17}}
\end{minipage}
\begin{minipage}[b]{0.50\linewidth}
\centering
\subfloat[CP and Tucker - 50MS]{
\includegraphics[scale=0.35]{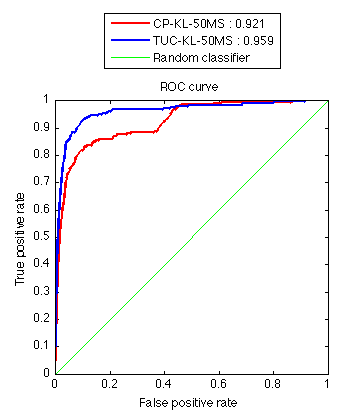}\label{fig:subfig18}}
\end{minipage}
\caption{Link prediction result with missing slices and KL cost}
\label{fig:missingSlice}
\vspace{-15pt}
\end{figure}

We also study the missing slice problem, which is particularly important in link prediction because we may often have new users starting to use an application, e.g., a location-activity recommender system. Since they are new users, they will have no entry in $X_{1}$, i.e., a completely missing slice. It is not possible to reconstruct a missing slice of a tensor using its low-rank approximation. A similar argument is valid in the case of matrices for completely missing rows/columns \cite{CaPl09}. In such cases, additional sources of information will be useful to make recommendations to new users. We observe that our coupled models could predict the links when there is no information about a user in tensor $X_{1}$, by utilizing the additional sources of information. We test this case by setting slices to missing randomly in $X_{1}$. Figure~\ref{fig:missingSlice} demonstrates the performance of coupled models with KL divergence when 10 users' data and 50 users' data are missing. Note that Tucker is superior to CP as the amount of missing data increases.

\section{Conclusions}
\label{sec:conc}

In this study, we have studied link prediction problem using coupled analysis of relational data represented as datasets in the form of matrices and higher-order tensors. The problem is formulated as simultaneous factorisation of higher-order tensors/matrices extracting common latent factors from the shared modes. While most existing studies on coupled analysis have been developed to fit a specific type of a tensor model using a particular loss function, we have used  Generalized Coupled Tensor Factorisation framework, which enables us to develop coupled models for joint analysis of multiple data sets using various tensor models and cost functions. In our coupled analysis for link prediction, we have studied both KL-divergence and Euclidean distance-based cost functions as well as different tensor models. Our numerical results on a real GPS data demonstrate that selection of the tensor model and the loss function is important in terms of link prediction performance. While our experiments have been limited to a dataset, which is not large-scale, the updates used in GCTF respect the sparsity pattern in the data; therefore, the proposed approach scales to large data. We plan to extend our study in that direction and show its applicability on large-scale data. 
\section{Acknowledgments}

This work is funded by the TUBITAK grant number 110E292, Bayesian matrix and tensor factorisations (BAYTEN) and Bo\u{g}azi\c ci University research fund BAP5723. It is also funded in part by the Danish Council for Independent Research | Technology and Production Sciences and Sapere Aude Program under the projects 11-116328 and 11-120947.

%
%
\bibliographystyle{splncs}
\bibliography{paper}

\end{document}